\documentclass[10pt,twocolumn,letterpaper]{article}

\usepackage{wacv}
\usepackage{times}
\usepackage{epsfig}
\usepackage{graphicx}
\usepackage{amsmath}
\usepackage{amssymb}
\usepackage{bm}
\usepackage{multirow}
\usepackage{subfig}
\usepackage{changes}
\usepackage[defernumbers, style=ieee]{biblatex}
\wacvfinalcopy 

\def\wacvPaperID{651} 

\makeatletter
\let\blx@rerun@biber\relax
\makeatother

\ifwacvfinal\fi
\DeclareMathOperator*{\argmin}{arg\,min}
\DeclareMathOperator*{\argmax}{arg\,max}
\begin{document}

\title{Spatial Knowledge Distillation to aid Visual Reasoning}

\author{Somak Aditya$\dagger$\footnotemark{}\\
Adobe Research (BEL)\\
{\tt\small saditya@adobe.com}
\and
Rudra Saha*, Yezhou Yang, Chitta Baral \\
CIDSE, Arizona State University\\
{\tt\small \{rsaha3,yz.yang,chitta\}@asu.edu}
}
\maketitle
\ifwacvfinal\thispagestyle{empty}\fi
\footnotetext{* denotes equal contribution. $\dagger$: work was done when author was in CIDSE, ASU.}
\vspace{-20pt}
\begin{abstract}

For tasks involving language and vision, the current state-of-the-art methods tend not to leverage any additional information that might be present to gather relevant (commonsense) knowledge. 
A representative task is Visual Question Answering where large diagnostic datasets have been proposed to test a system's capability of answering questions about images. 
The training data is often accompanied by annotations of individual object properties and spatial locations.
In this work, we take a step towards integrating this additional privileged information in the form of spatial knowledge to aid in visual reasoning. We propose a framework that combines recent advances in knowledge distillation (teacher-student framework), relational reasoning and probabilistic logical languages to incorporate such knowledge in existing neural networks for the task of Visual Question Answering. Specifically, for a question posed against an image, we use a probabilistic logical language to encode the spatial knowledge and the spatial understanding about the question in the form of a mask that is directly provided to the teacher network. The student network learns from the ground-truth information as well as the teacher’s prediction via distillation. We also demonstrate the impact of predicting such a mask inside the teacher’s network using attention. Empirically, we show that both the methods improve the test accuracy over a state-of-the-art approach on a publicly available dataset.

\end{abstract}

\vspace{-15pt}

\section{Introduction}
Vision and language tasks such as Visual Question Answering (VQA)
are often considered as ``AI-complete'' tasks \cite{antol2015vqa} since they require multi-modal knowledge beyond a single sub-domain. Recently, the VQA1.0 dataset was proposed as a representative dataset for the task of VQA \cite{antol2015vqa}. This task aims to combine efforts from three broad sub-fields of AI namely image understanding, language understanding and reasoning. Despite its popularity, most of its questions focus on object recognition in images and natural language understanding. Question-Image pairs where a system may require compositional reasoning or reasoning with external knowledge, seem to be largely absent. 
To explicitly assess the reasoning capability, several specialized datasets have been proposed, that emphasize specifically on questions requiring complex multiple-step reasoning (CLEVR \cite{johnson2016clevr}, Sort-of-Clevr \cite{santoro2017simple}) or questions that require reasoning using external knowledge (F-VQA \cite{wang2017fvqa}). 

Current state-of-the-art neural architectures do not explicitly model such external knowledge and reasoning with them to solve visual reasoning tasks. Several researchers \cite{lake2016building, yannPathAI} in their works have pointed out the necessity of explicit modeling of such knowledge. This 
necessitates  considering the following issues:

i) \textit{What kind of knowledge is needed}? ii) \textit{Where and how to get them}? and iii) \textit{What kind of reasoning mechanism to adopt for such knowledge}?


\begin{figure}[!htb]
	\centering
    \includegraphics[width=0.5\textwidth]{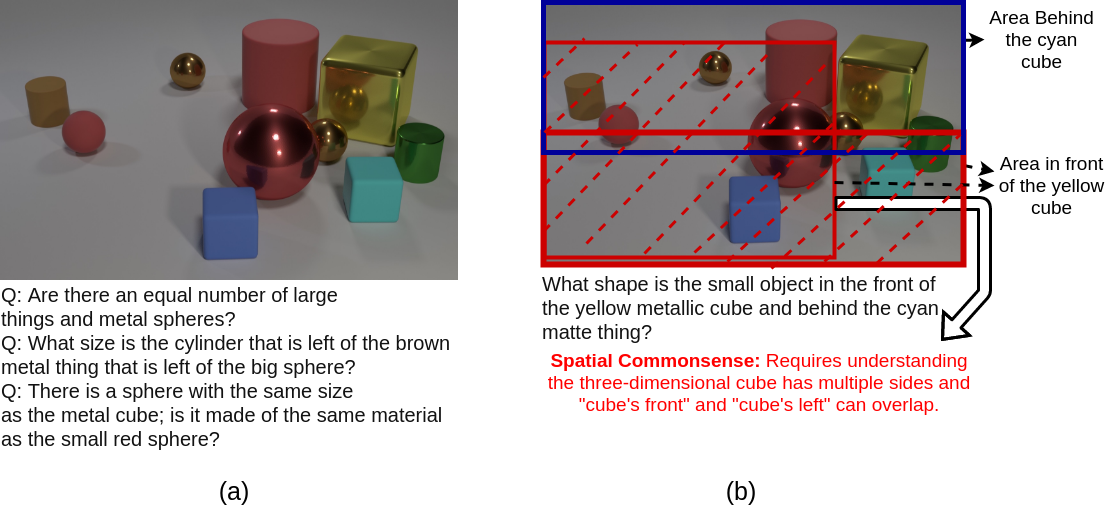} 
    \caption{(a) An image and a set of questions from the CLEVR dataset. Questions often require multiple-step reasoning, for example \textit{in the second question, one needs to identify the big sphere, then  recognize the reference to the brown metal cube, which then refers to the root object, that is, the brown cylinder. (b) An example of spatial commonsense knowledge needed to solve a CLEVR-type question.}}
    \label{clevr1}
\end{figure}

To understand the kind of external knowledge required, we investigate the CLEVR dataset proposed in \cite{johnson2016clevr}. This dataset explicitly asks questions that require relational and multi-step reasoning. An example is provided in Fig.~\ref{clevr1}(a). In this dataset, the authors create synthetic images consisting of a set of objects that are placed randomly within the image. Each object is 
created randomly by varying its shape, color, size and texture. For each image, $10$ complex questions are generated. Each question inquires about an object or a set of objects in the image. To understand which object(s) the question is referring to, one needs to decipher the clues that are provided about the property of the object or the spatial relationships with other objects. This can be a multiple-step process, that is: first recognize object A, that refers to object B, which refers to C and so on. There have been multiple architectures proposed to answer such complex questions. Authors in \cite{hu2017learning} attempt to learn a structured program from the natural language question. This program acts as a structured query over the objects and relationship information provided as a scene graph and can retrieve the desired answer. More interestingly, the authors in \cite{santoro2017simple} model relational reasoning explicitly in the neural network architecture and propose a generic relational reasoning module to answer questions. This is one of the first known attempt to formulate a differentiable function to embody a generic relational reasoning module that is traditionally formulated using logical reasoning languages. The failure cases depicted by this work, often points to the lack of complex commonsense knowledge such as, \textit{the front of cube should consist of front of all visible side of cubes.} These examples point that spatial commonsense knowledge might help answer questions such as in Fig.~\ref{clevr1}(b). Even though procuring such knowledge explicitly is difficult, we observe that parsing the questions and additional scene-graph information can help ``disambiguate'' the area of the image on which a phrase of a question focuses on.


In this work, to the best of our knowledge, we 
{\color{black}{make the first attempt }}
of incorporating such additional {\color{black}{information in the form of}} spatial knowledge {\color{black}{into existing architectures for}} vision and language tasks. Specifically, we concentrate on the task of VQA for questions that require multiple-step (relational) reasoning, and we explore how a recently proposed relational reasoning based architecture \cite{santoro2017simple} can be improved further with the aid of additional spatial knowledge extracted from the image using the question and the scene-graph information. This is an important avenue, as humans often use a large amount of external knowledge to solve tasks that they have acquired through years of experience\footnote{The authors in \cite{lake2016building} quoted a reviewer's comment: ``Human learners - unlike DQN and many other deep Learning systems - approach new problems armed with extensive prior experience.''. The authors also ask ``How do we bring to bear rich prior knowledge to learn new tasks and solve new problems?''.}. To extract such spatial knowledge in a form that can be integrated and reasoned with, we take inspiration from techniques from the field of Knowledge Representation and Reasoning (KR\&R) and utilize a reasoning engine viz. Probabilistic Soft Logic (PSL) \cite{kimmig:probprog12}. In practice, these languages and their available implementations are often susceptible to the high amount of noises in real-world datasets and hence, their direct applications have been somewhat limited. One can assume that in order to provide robust, interpretable and accurate solutions, one needs to leverage both the robustness and interpretability of declarative logical reasoning languages and the high-level representation learning capability of deep learning.

We rely on the knowledge distillation paradigm \cite{hinton14071698} in order to integrate the extracted spatial knowledge with the relational reasoning architecture. 
{\color{black}{Knowledge distillation}} aims to transfer the predictions learned by a complex model, often regarded as a teacher model, to that of a simpler model, usually deemed as a student model, via distillation \cite{hinton14071698}. 
There are various flavours of the knowledge distillation paradigm depending upon the complexity of the teacher's model as well as the amount of information it has access to relative to the student's model.

Thus to solve our task, we propose a student-teacher network based architecture, where the teacher has access to privileged information. For the VQA, this priveledged information is the spatial knowledge required to answer the question in the form of an attentive image mask based on the question and the scene-graph information. The student network is the existing architecture we want to distill this knowledge into. 
We provide two methods for calculating the mask; i) when the object and relationships are provided for an image, one can calculate a mask using probabilistic reasoning, and ii) if such data is not available, such a mask can be calculated inside the network using attention. We experiment on the Sort-of-Clevr dataset, and empirically show that both these methods outperform a state-of-the-art relational reasoning architecture. We observe that the teacher model (using the spatial knowledge inferred by PSL inference) achieves a sharp {\color{black}{$13.7\%$}} jump in test accuracy over the baseline architecture. {\color{black}{The existing architecture \ie the student model, distilled with this knowledge, shows a generalization boost of $6.2\%$ as well}}. We also provide ablation studies of the reasoning mechanism on (questions and scene information from) the CLEVR dataset.
\section{Related Work}
 Our work is influenced by the following thrusts of work: probabilistic logical reasoning, spatial reasoning, reasoning in neural networks,  knowledge distillation; and the target application area of Visual Question Answering.

Recently, researchers from the KR$\&$R community, and the Probabilistic Reasoning community have come up with several robust probabilistic reasoning languages which are deemed more suitable to reason with real-world noisy data, and incomplete or noisy background knowledge. Some of the popular ones among these reasoning languages are Markov Logic Network \cite{Richardson:2006:MLN:1113907.1113910}, Probabilistic Soft Logic \cite{kimmig:probprog12},  and ProbLog \cite{DeRaedt:2007:PPP:1625275.1625673}. Even though these new theories are considerable large steps towards modeling uncertainty (beyond previous languages engines such as Answer Set Programming \cite{baral2003knowledge}); the benefit of using these reasoning engines has not been successfully shown on large real-world datasets. This is one of the reasons, recent advances in deep learning, especially the works of modeling knowledge distillation \cite{hinton14071698,vapnik2015learning} and relational reasoning have received significant interest from the community. 
   
Modeling of Spatial Knowledge and reasoning using such knowledge in 2D or 3D space has given rise to multiple interesting works in both Computer Vision and Robotics, collectively termed as Qualitative Spatial Reasoning (QSR). {\color{black}{Randell \etal}} \cite{Randell92aspatial} proposed an interval logic for reasoning about space. {\color{black}{Cohn and Renz}} \cite{COHN2008551} proposed advancements over previous languages aimed at robotic navigation in 2D or 3D space. In these languages, the relations between two objects are modeled spatially.
Our work is also influenced by this series of works (such as Region Connection Calculus etc.), in the sense of what ``privileged information'' we expect along-with the image and the question. For the CLEVR dataset, the relations \texttt{left, right, front, behind} can be used as a closed set of spatial relations among the objects and that often suffices to answer most questions. For real images, a scene graph that encodes spatial relations among objects and regions, such as proposed in \cite{elliott2013image} 
would be useful to integrate our methods.
   
Popular probabilistic reasoning mechanisms from the statistical community often define distribution with respect to Probabilistic Graphical Models. There have been a few attempts to model such graphical models in conjunction with deep learning architectures \cite{Zheng:2015:CRF:2919332.2919659}. However, multi-step relational reasoning, and reasoning with external domain or commonsense knowledge\footnote{An example of multi-step reasoning:  if event $A$ happens, then $B$ will happen. The event $B$ causes action $C$ only if event $D$ does not happen. For reasoning with knowledge: consider for a image with a giraffe, we need to answer ``Is the species of the animal in the image and an elephant same?''} require the robust structured modeling of the world as adopted by KR$\&$R languages. In its popular form, these reasoning languages often use 
predicates to describe the current world, such as $color(hair, red)$, $shape(object_1,sphere)$, $material(object_1,metal)$ etc; and then declare rules that the world should satisfy.  Using these rules, truth values of unknown predicates are obtained, such as $ans(?x,O)$ etc. Similarly, the work in \cite{santoro2017simple}, defines the relational reasoning module as $ RN(O) = f_{\phi} \Big(\sum_{i,j} g_{\theta}(o_i,o_j)\Big)$, 
where $O$ denote all objects. In this work, the relation between a pair of objects  (i.e. $g_\theta$) and the final function over this collection of relationships i.e. $f_\phi$ are defined as multilayer perceptrons (MLP) and are learnt using gradient descent in an end-to-end manner. This model's simplicity and its close resemblance to traditional reasoning mechanisms motivates us to pursue further and integrate external knowledge.
 
Several methods have been proposed to distill knowledge from a larger model to a smaller model or from a model with access to privileged information to a model without such information. Hinton \etal \cite{hinton14071698} first proposed a framework where a large cumbersome model is trained separately and a smaller student network learns from both groundtruth labels and the large network. Independently, Vapnik \etal. \cite{vapnik2015learning} proposed an architecture where the larger (or the teacher) model has access to privileged information and the student model does not. These models together motivated many natural language processing researchers to formulate textual classification tasks as a teacher-student model, where the teacher has privileged information, such as a set of rules; and the student learns from the teacher and the ground-truth data. The imitation parameter controls how much the student \textit{trusts} the teacher's decision. In \cite{hu-EtAl:2016:EMNLP2016}, an iterative knowledge distillation is proposed where the teacher and the student learn iteratively and the convolutional network's parameters are shared between the models. In \cite{hu-EtAl:2016:P16-1}, the authors propose to solve sentiment classification, by encoding explicit logical rules and integrating the grounded rules with the teacher network. These applications of teacher-student network only exhibited success with classification problems with very small number of classes (less than three). 
 
In this paper, we show a knowledge distillation integration with privileged information which is applied to a $28$-class classification, and we observe that it improves by a large margin on the baseline. In \cite{yu2017_vrd_knowledge_distillation}, the authors use encoded linguistic knowledge in the form of $P(pred|obj,subj)$ to perform Visual Relationship Detection. In this work, we apply knowledge distillation in a visual question answering setting, that require both visual reasoning and question understanding. 
 
In the absence of the scene information or in cases where such information is expensive to obtain, an attention mask over the image can be predicted inside the network based upon the posed question. Attention mechanism has been successfully applied in image captioning \cite{xu2015show,mun2017text}, machine translation\cite{bahdanau2014neural,vaswani2017attention} and visual question answering \cite{yang2016stacked}. In \cite{yang2016stacked}, a stacked attention network was used to predict a mask over the image. They use the question vector separately to query specific image features to create the first level of attention. In contrast, we combine the question vector with the whole image features to predict a coarse attention mask.

\begin{figure*}[htb!]
	\centering
    \includegraphics[width=0.9\textwidth,height=0.25\textheight]{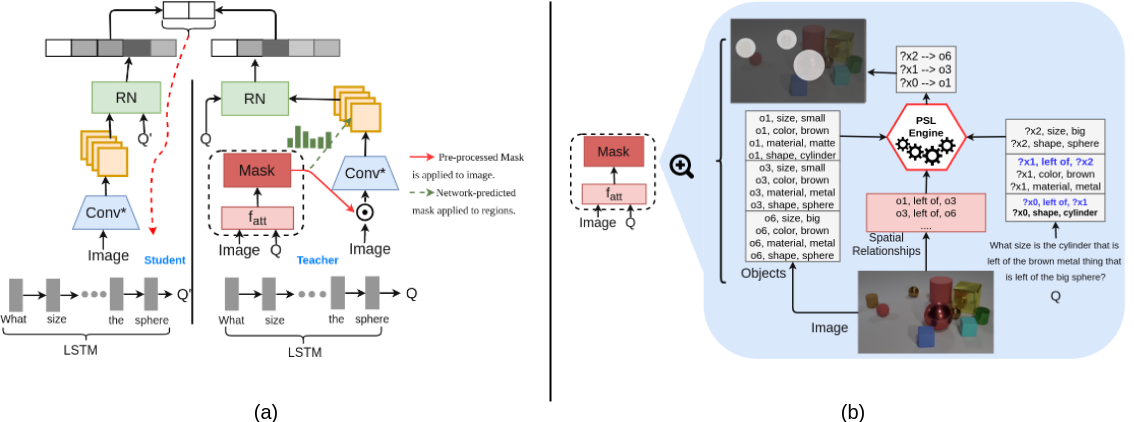}
    \caption{(a) The Teacher-Student Distillation Architecture. As the base of both teacher and student, we use the architecture proposed by the authors in \cite{santoro2017simple}. For the experiment with pre-processed mask generation, we pass a masked image through the convolutional network and for the network-predicted mask, we use the image and question to predict an attention mask over the regions. (b) We show the internal process of mask creation. }
    \label{archi}
    \vspace{-10pt}
\end{figure*}
\section{Additional Knowledge Integration Method}
In this section, we explain the various components of our proposed framework for integrating additional spatial information with existing neural architectures. We start by formalizing the probabilistic reasoning mechanism which enables us to extract such spatial knowledge in the presence of scene information. Then, we describe the knowledge distillation paradigm \cite{hinton14071698} that enables us to infuse this extracted knowledge into existing networks which in our case is a relational reasoning architecture \cite{santoro2017simple}. We also outline the in-network computation required in the absence of the scene-graph information. 

\subsection{Probabilistic Reasoning Mechanism}
In order to reason about the spatial relations among the objects in a scene and textual mentions of those objects in the question, we choose Probabilistic Soft Logic (PSL) \cite{kimmig:probprog12} as our reasoning engine. Using PSL provides us three  advantages: i) (Robust Joint Modeling) from the statistical side, PSL models the joint distribution of the random variables using a Hinge-Loss Markov Random Field, ii) (interpretability) we can use clear readable declarative rules that (directly) relates to defining the clique potentials, and iii) (Convex Optimization) the optimization function of PSL is designed in a way so that the underlying function remains convex and that provides an added advantage of faster inference. We use PSL, as it has been successfully used in Vision applications  \cite{london2013collective} in the past and it is also known to scale up better than its counterparts \cite{Richardson:2006:MLN:1113907.1113910}.

\vspace{-6pt}

\subsubsection{Hinge-Loss Markov Random Field and PSL}   

Hinge-Loss Markov Random Fields (HL-MRF)  is a general class of continuous-valued  probabilistic graphical model. An HL-MRF is defined as follows:
Let $\bm{y}$ and $\bm{x}$ be two vectors of $n$ and ${n'}$ random variables respectively, over the domain $D =[0,1]^{n+{n'}}$. The feasible set $\tilde{D}$ is a subset of $D$, which satisfies a set of inequality constraints over the random variables.

A \textit{Hinge-Loss Markov Random Field} $\mathbb{P}$ is a probability density over $D$, defined as: if $(\bm{y},\bm{x}) \notin \tilde{D}$, then $\mathbb{P}(\bm{y}|\bm{x})=0$; if $(\bm{y},\bm{x}) \in \tilde{D}$, then:
\begin{equation}
\small
\begin{aligned}
\mathbb{P}(\bm{y}|\bm{x}) \propto  exp(-f_{\bm{w}}(\bm{y},\bm{x})).
\end{aligned}
\end{equation}

PSL combines the declarative aspect of reasoning languages with conditional dependency modeling power of undirected graphical models. In PSL a set of weighted if-then rules over first-order predicates is used to specify a Hinge-Loss Markov Random field. 

 

  In general,  let $\bm{C}=(C_1,...,C_m)$ be such a collection of weighted rules where each $C_j$ is a disjunction of literals, where each literal is a variable $y_i$ or its negation $\neg y_i$, where $y_i\in \bm{y}$. Let $I_j^{+}$ (resp. $I_j^{-}$ ) be
  the set of indices of the variables that are not negated (resp. negated) in $C_j$. Each $C_j$ can be represented as: 
\begin{equation}
\setlength{\abovedisplayskip}{1pt}
\setlength{\belowdisplayskip}{1pt}
w_j : \lor_{i \in I_j^{+}} y_i \leftarrow \land_{i \in I_j^{-}} y_i ,
\end{equation}

or equivalently, $w_j: \lor_{i \in I_j^{-}} (\neg y_i) \bigvee \lor_{i \in I_j^{+}} y_i$. 
  A rule $C_j$ is associated with a non-negative weight $w_j$. PSL relaxes the boolean truth values of each ground atom $a$ (constant term or predicate with all variables replaced by constants) to the interval [0, 1], denoted as $V(a)$. To compute soft truth
values, 
Lukasiewicz's relaxation \cite{klir1995fuzzy} of conjunctions ($\land$), disjunctions ($\lor$) and negations ($\neg$) 
are used:
\begin{equation*}
\small
\begin{aligned}
V(l_1\land l_2) = max\{0,V(l_1)+V(l_2)-1\}\\
V(l_1\lor l_2) = min\{1,V(l_1)+V(l_2)\}\\
V(\neg l_1) = 1-V(l_1).
\end{aligned}
\end{equation*}

In PSL, the ground atoms are considered as random variables, and the joint distribution is modeled using Hinge-Loss Markov Random Field (HL-MRF).

  In PSL, the hinge-loss energy function $f_{\bm{w}}$ is defined as:
\begin{equation}
\small
f_{\bm{w}}(\bm{y}) = \sum\limits_{C_j \in \bm{C}} w_j\text{ }max\big\{ 1- \sum_{i \in I_j^{+}} V(y_i) - \sum_{i \in I_j^{-}} (1- V(y_i)),0\big\}.
\end{equation} 
The maximum-a posteriori (MAP) inference objective of PSL becomes:
{\small
\begin{equation}
\begin{aligned}
\setlength{\abovedisplayskip}{0pt}
\argmax_{\bm{y}\in [0,1]^n} P(\bm{y}) &\equiv \argmax_{\bm{y}\in [0,1]^n} exp(-f_{\bm{w}}(\bm{y})) \\
 &\equiv \argmin_{\bm{y}\in [0,1]^n} \sum\limits_{C_j \in \bm{C}} w_j\text{ }max\Big\{ 1- \sum_{i \in I_j^{+}} V(y_i) \\
&- \sum_{i \in I_j^{-}} (1-V(y_i)),0\Big\},
\label{eq:4}
\end{aligned}
\end{equation}
}
  where the term $w_j\times max\big\{ 1- \sum_{i \in I_j^{+}} V(y_i) - \sum_{i \in I_j^{-}} (1- V(y_i)),0\big\}$ measures the ``distance to satisfaction'' for each grounded rule $C_j$.

\subsection{Knowledge Distillation Framework}

While PSL provides a probabilistic knowledge representation {\color{black}{, as shown in Fig.~\ref{archi}(b)}}, a mechanism is needed to utilize them under the deep neural networks based systems.
We use the generalized knowledge distillation paradigm \cite{lopezpaz2015unifying}, where the teacher's network can be a larger network performing additional computation or have access to privileged information, to achieve this integration resulting in two different architectures i) (External Mask) teacher with provided ground-truth mask, ii) (In-Network Mask) teacher predicts the mask with additional computation. Here, we provide general formulations for both methods {\color{black}{and give an overview of how the external mask is calculated}}\footnote{\color{black}{A detailed example of how we estimate these predicates is provided in Supplementary material.}}.

\vspace{-5pt}

\subsubsection{General Architecture}

The general architecture for the teacher-student network is provided in Fig.~\ref{archi}(a). Let us denote the teacher network as $\bm{q}_\phi$ and the student network as $\bm{p}_\theta$. In both scenarios, the student network uses the relational reasoning network \cite{santoro2017simple} to predict the answer.  The teacher network uses an LSTM to process the question, and a convolutional neural network to process the image. Features from the convolutional network and the final output from the LSTM is used as input to the relational reasoning module to predict an answer. Additionally in the teacher network, we predict a mask. For the External Mask setting, the mask is predicted by a reasoning engine and applied to the image, and for the attention setting, the mask is predicted using the image and text features and applied over the output from the convolution. The teacher network $\bm{q}_\phi$ is trained using softmax cross-entropy loss against the ground truth answers for each question. The student network is trained using knowledge distillation with the following objective:

\vspace{-0.4cm}

\begin{equation}
\setlength{\abovedisplayskip}{1pt}
\setlength{\belowdisplayskip}{1pt}
\begin{aligned}
\theta = \argmin_{\theta \in \Theta} \sum\limits_{n=1}^{N} &(1-\pi) \ell_1(\bm{y_n}, \sigma_\theta(\bm{x_n})) \\& + \pi \ell_2(\bm{s_n}, \sigma_\theta(\bm{x_n})),
\end{aligned}
\label{eqpsl}
\end{equation} 

\vspace{-0.2cm}

where $\bm{x_n}$ is the image-question pair, and $\bm{y_n}$ is the answer that is available during the training phase; the $\sigma_\theta(.)$ is the usual \textit{softmax} function; $\bm{s_n}$ is the soft prediction vector of $\bm{q}_\phi$ on $\bm{x_n}$ and $\ell_i$ denotes the loss functions selected according to specific experiments (usually $\ell_1$ is cross-entropy and $\ell_2$ is euclidean norm). $\pi$ is often called the imitation parameter and determines how much the student trusts the teacher's predictions.
\subsubsection{External Mask Prediction}
\label{sec:ex_mask}
\vspace{-5pt}
This experimental setting is motivated by the widely available scene graph information in large datasets starting from Sort-of-Clevr and CLEVR to Visual Genome. We use the following information about the objects and their relationships in the image: i) the list of \textit{attribute, value} pairs for each object, ii) the spatial relationships between objects, and iii) each object's relative location in the image. 
   
We view the problem as a special case of the bipartite matching problem, where there is one set of textual mentions ($M$) of the actual objects and a second set of actual objects ($O$). Using probabilistic reasoning we find a matching between object-mention pairs based on how the attribute-value pairs match between the objects and the corresponding mentions, and when mention-pairs are consistently related (such as \textit{larger than, left to, next to}) as their matched object-pairs. Using the scene graph data, and by parsing the natural language question, we estimate the value of the following predicates: $attr_o(O,A,V)$, $attr_m(M,A,V)$ and $consistent(A,O,O_1,M,M_1)$. The predicate $attr_m(M,A,V)$ denotes the confidence that the value of the attribute $A$ of the textual mention $M$ is $V$. The predicate $attr_o(O,A,V)$  is similar and denotes a similar confidence for the object $O$. The predicate $consistent(R,O,O_1,M,M_1)$ indicates the confidence that the textual mentions $M$ and $M_1$ are consistent based on a relationship $R$ (spatial or attribute based), if $M$ is identified with the object $O$ and $M_1$ is identified with the object $O_1$. Using only these two predicate values, we use the following two rules to estimate which objects relate to which textual mentions.
\begin{equation*}
\setlength{\abovedisplayskip}{1pt}
\setlength{\belowdisplayskip}{1pt}
\begin{aligned}
w_1: candidate(M,O) &\leftarrow object(O) \land mention(M) \\&\land attr_o(O,A,V) \land attr_m(M,A,V).
\end{aligned}
\end{equation*}  
\begin{equation*}
\begin{aligned}
w_2: candidate(M,O) &\leftarrow object(O) \land mention(M) \\&\land candidate(M,O) \\&\land candidate(M_1,O_1) \\& \land consistent(A,O,O_1,M,M_1).
\end{aligned}
\end{equation*}  
We use the grounded rules (variables replaced by constants) to define the clique potentials and use eq.~\ref{eq:4} to find the confidence scores of grounded $candidate(M,O)$ predicates. Using this mention to object mapping, we use the objects that the question refers to. For each object, we use the center location, and create a heatmap that decays with distance from the center. We use a union of these heatmaps as the mask. This results into a set of spherical masks over the objects mentioned in the question, as shown in Fig.~\ref{archi}(b). To validate our calculated masks, we annotate the CLEVR validation set with the ground-truth objects, using the ground-truth structured program. We observe that our PSL-based method can achieve a $75\%$ recall and $70\%$ precision in predicting the ground-truth objects for a question.

\begin{figure}[!htb]
	\centering
    \includegraphics[height=0.15\textheight]{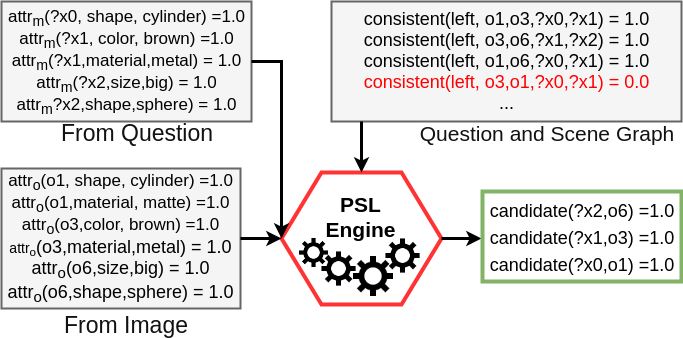}
    \caption{We elaborate on the calculated PSL predicates for the example image and question in Figure \ref{archi}(b). The underlying optimization benefits from the negative examples (the \textit{consistent} predicate with $0.0$, marked in red). Hence, these predicates are also included in the program.}
    \label{fig:example_psl}
\end{figure}

In Figure \ref{fig:example_psl}, we provide more details of the calculated PSL predicates for the example question and image in Figure \ref{archi}(b). We use this top collection of objects and their relative locations to create small spherical masks over the relevant objects in the images. 
 
\vspace{-20pt}

\subsubsection{In-Network Mask Prediction}

The External Mask setting requires privileged information such as scene graph data about the image, which includes the spatial relations between objects. Such information is often expensive to obtain. Hence, in one of our experiments, we attempt to emulate the mask creation inside the network. We formulate the problem as attention mask generation over image regions using the image ($\bm{x_I} \in \mathbb{R}^{64\times64\times3}$) and the question ($\bm{x_q} \in \mathbb{R}^{w\times d}$). The calculation can be summarized by the following equations:
    
\begin{equation}
\begin{aligned}
r_I &= conv^{*}(\bm{x_I}). ~~
q_{emb} = LSTM(\bm{x_q}).\\
v &= tanh(W_I r_I + W_q q_{emb} + b).\\
\alpha &= \exp(v)/\sum\limits_{r=1}^{x*y} \exp(v_r),
\end{aligned}
\end{equation}    
   
 where $r_I$ is $x \times y$ regions with $o_c$ output channels, $q_{emb} \in \mathbb{R}^{h}$ is the final hidden state output from $LSTM$ (hidden state size is $h$); $W_I (\in \mathbb{R}^{xyo_c \times xy})$ and $W_q (\in \mathbb{R}^{xy \times h})$ are the weights and $b$ is the corresponding bias. Finally, the attention $\alpha$ over regions is obtained by exponentiating the weights and then normalizing them. The attention $\alpha$ is then reshaped and element-wise multiplied with the region features extracted from the image. This is considered as a mask over the image regions conditioned on the question vector and the image features.

\section{Experiments and Results}
\label{sec:exp}

We propose two architectures, one where the teacher has privileged information and the other where the teacher performs additional calculation using {\color{black}{auxiliary}} in-network modules. We perform experiments to validate whether the direct addition of information (external mask), or additional modules (model with attention) improves the teacher's performance over the baseline. We also perform similar experiments {\color{black}{to validate whether this learned knowledge can be distilled to existing neural networks (student model)}} 
. Additionally, we conduct ablation studies on the probabilistic logical mechanism using which we predict a ground-truth mask from the question and the scene information.

\subsection{Setup}
As our testbed, we use the ``Sort-of-Clevr'' from \cite{santoro2017simple} and the CLEVR dataset from \cite{johnson2016clevr}. As the original Sort-of-Clevr dataset is not publicly available,  we create the synthetic dataset as described by the authors\footnote{We make the code and data available in supplementary material.}. We use similar specification, i.e., there are $6$ objects per image, where each object is either a circle or a rectangle, and we use $6$ colors to identify each different object. Unlike the original dataset, we generate natural language questions along with their one-hot vector representation. In our experiments we primarily use the natural-language question. We only use the one-hot vector to replicate results of the baseline Relational Network (RN)\footnote{We were unable to replicate the results of \cite{santoro2017simple} on CLEVR dataset. This is why we used another baseline (Stacked Attention Network) and show how our method improves on that baseline. The primary reason being the original network was trained by authors on 10 parallel GPUs on $640$ batch size. This was not feasible to replicate in lab setting. Based on our experiments, the best accuracy obtained by the relational reasoning network is $68\%$ with a batch-size of $640$ on a single-GPU worker, after running for $600$ epochs over the dataset.}. For our experiments, we use $9800$ images for training, $200$ images each for validation and testing. There are $10$ question-answer pairs for each image. For Sort-of-Clevr, we use four convolutional layers with 32, 64, 128 and 256 kernels, ReLU non-linearities, and batch normalization. The questions were passed through an LSTM where the word embeddings are initialized with 50-dimensional Glove embeddings \cite{pennington2014glove}. The LSTM output and the convolutional features are passed through the RN network\footnote{A four-layer MLP consisting of $2000$ units per layer with ReLU non-linearities is used for $g_\theta$; and a four-layer MLP consisting of 2000, 1000, 500, and 100 units with ReLU non-linearities used for $f_\phi$.}.  The baseline model was optimized with a cross-entropy loss function using the Adam optimizer with a learning rate of $1e^{-4}$ and mini-batches of size $64$. For CLEVR, we use the Stacked Attention Network \cite{yang2016stacked} with the similar convolutional network and LSTM as above. We get similar results with VGG-16 as the convolutional network. Instead of the RN layer, we pass the two outputs through two levels of stacked attention, followed by a fully-connected layer.
On top of this basic architecture, we define the student and teacher networks. The student network uses the same architecture as the baseline. We propose two variations of the teacher network, and we empirically show how these proposed changes improve upon the performance of the baseline network.

\subsection{External Mask Prediction}
\begin{figure}[!htb]
    \centering
    {\includegraphics[width=0.48\textwidth, height=0.16\textheight]{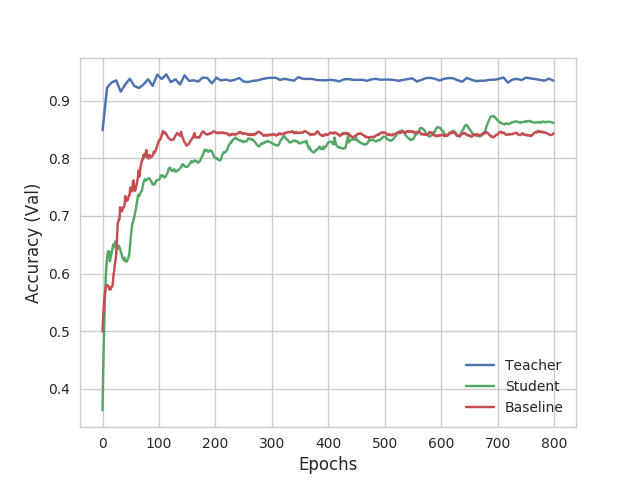}}
    \caption{Validation accuracy after each epoch for teacher and student networks for iterative knowledge distillation on Sort-of-Clevr dataset and compare with the baseline.}
    \label{fig:e3_student_acc_b}
\end{figure}

\begin{figure}[!htb]
    \centering
    {\includegraphics[width=0.42\textwidth]{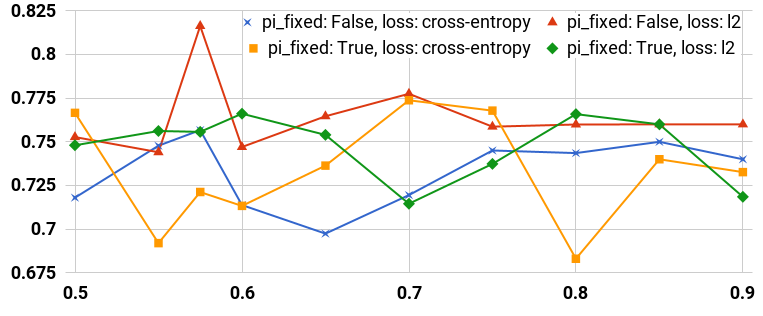}}
    \caption{External Mask Prediction: Test Accuracy for different hyperparamter combination to obtain the best imitation parameter ($\pi$) for student in sequential distillation.}
    \label{fig:e3_student_acc_a}
\end{figure}

In this setting of the experiment, the ground-truth mask, as calculated in ~\ref{sec:ex_mask}, is element-wise multiplied to the image and then the image is passed through the convolutional network. We experiment with both sequential and iterative knowledge distillation. In the sequential setting, we first train the teacher network for $100$ epochs with random embedding size of $32$, batch size as $64$, learning rate $0.0001$. In the previous attempts to use distillation in natural language processing (\cite{hu-EtAl:2016:P16-1,kim2016sequence}), the optimal value of $\pi$ has been reported as $\min(0.9,1-0.9^t)$ or $0.9^t$. Intuitively, either at the early or at the latter stages, the student almost completely \textit{trusts} the teacher. However, our experiments show different results. For the student network, we employ a hyperparamter search on the value of imitation parameter $\pi$ and use two settings, where $\pi$ is fixed throughout the training and in the second setting, $\pi$ is varied using $\min(\pi,1-\pi^t)$. We vary the loss $\ell_2$ among cross entropy and euclidean norm. The results of the hyperparameter optimization experiment is depicted in Fig.~\ref{fig:e3_student_acc_a}. From this experiment, it can be observed that varying $\pi$ over epochs gives better results than using a fixed $\pi$ value for training the student. We observe a sharp increase in accuracy using the $\pi$ value $0.575$. This result is more consistent with the parameter value chosen by the authors in \cite{yu2017_vrd_knowledge_distillation}. We also experiment by varying the word embedding ($50$-dimensional glove embedding and $32$-dimensional word embedding) and learning rate. For sequential knowledge distillation, we get the best results with glove embedding and learning rate as $1e^{-4}$. However, we get huge improvements by using iterative knowledge distillation, where in each alternate epoch the student learns from the teacher and the groundtruth data; and the teacher learns from its original loss function and the student's soft prediction (similar to Eqn.~\ref{eqpsl}). Both weighted loss functions use the imitation parameter $0.9$ (which remains fixed during training). {\color{black} We show the gradual learning of the teacher and the student till 800 epochs in Fig.~\ref{fig:e3_student_acc_b}
and compare it with the RN baseline.  We observe that: 1) the External Mask-augmented Teacher network converges faster than the baseline and 2) the Student network outperforms the baseline after $~650$ epochs of training}.

\subsection{Larger Model with Attention}
\begin{table*}[!htb]
\centering
\begin{tabular}{c|c|cc|cc|cc}
              & \multirow{2}{*}{Baseline} &  \multicolumn{2}{c|}{External Mask} & \multicolumn{2}{c|}{In-Network Mask} & \multicolumn{2}{c}{Performance Boost Over Baseline ($\Delta$)}  \\
              &                           &  Teacher     & Student    & Teacher               & Student    & Teacher     & Student  \\\hline
Sort-of-Clevr &   $82$\% (\cite{santoro2017simple})   & $\boldsymbol{95.7}$\%            & $88.2$\%           &  87.5\%                     &        $82.8$\%        & $13.7$\%  & $6.2\%$   \\
CLEVR &   $53$\% (\cite{yang2016stacked})   &  $\boldsymbol{58}$\%        &     $55$\%       &        -               &   -             & $5$\%  & $2\%$    \\\hline
\end{tabular}
\vspace{5pt}
\caption{Test set accuracies of different architectures for the Sort-of-Clevr (with natural language questions) and CLEVR dataset. For CLEVR, we used the Stacked Attention Network (SAN) \cite{yang2016stacked} as baseline and only conducted the external-mask setting experiment as it already calculates in-network attention. Our re-implementation of SAN achieves $53$\% accuracy on CLEVR. Accuracy reported by \cite{santoro2017simple} on SAN is $61$\%. The reported best accuracy for Sort-of-Clevr and CLEVR are $94$\% (one-hot questions \cite{santoro2017simple}) and $97.8$\% (\cite{perez2017film}). 
}
\label{tab:acc}
\vspace{-15pt}
\end{table*}

\begin{figure}[!htb]
    \centering
    {\includegraphics[width=0.48\textwidth]{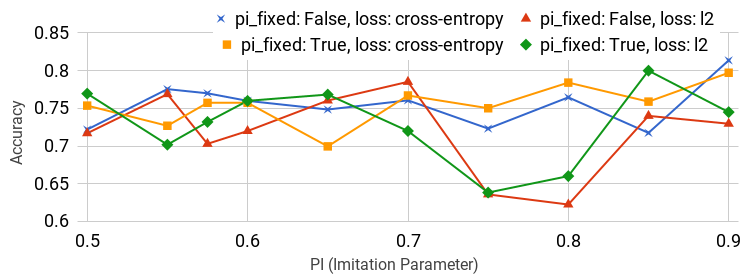}}
    \caption{ Model with Attention Mask: Test accuracy for the student network for different hyperparamter combination to obtain the best imitation parameter ($\pi$). We get the best validation accuracy using the $\pi$ as $0.9$, $\ell_2$ as cross entropy loss and varying $\pi$ by over epochs.}
    \label{fig:combined_acc2}
\end{figure}

In this framework, we investigate whether the mask can be learnt inside the network with attention mechanism.  We train the teacher network for $200$ epochs with glove vectors of size $50$, batch size as $64$, learning rate as $0.0001$. We have employed a hyperparamter search over learning rate, embedding type, and learning rate decay, and found that the above configuration produces best results. For the student network, we employed a similar hyperparamter search on the value of imitation parameter $\pi$ and use two settings, where $\pi$ is fixed throughout the training and in the second setting, $\pi$ is varied using $min(\pi,1-\pi^t)$. We also vary the learning rate  and the type of embedding (random with size 32 or glove vectors of size 50). The effect of the hyperparameter search is plotted in Fig.~\ref{fig:combined_acc2}. We have experimented with iterative knowledge distillation and the best accuracy obtained for the teacher and the student networks are similar to that of sequential setting. The best test accuracies of the student network, the teacher with larger model and the baselines are provided in Table \ref{tab:acc}.

\subsection{Analysis}
 The reported baseline accuracy on Sort-of-Clevr by \cite{santoro2017simple} is $94\%$ for both relational and non-relational questions. However, we use LSTMs to embed the natural language questions. Our implementation of the baseline achieves an overall test accuracy of $89\%$ with one-hot question representation and $82\%$ with LSTM embedding of the question. Addition of the pre-processed mask provides an increase in test accuracy to \textbf{95.7}$\%$. In contrast, the teacher model with attention mask achieves \textbf{87.5}$\%$. This is expected as the mask on the image simplifies the task by eliminating irrelevant region of image with respect to the question. 
 
\begin{figure}[!htb]
	\centering
    \subfloat{\includegraphics[width=0.5\textwidth, height=0.1\textheight]{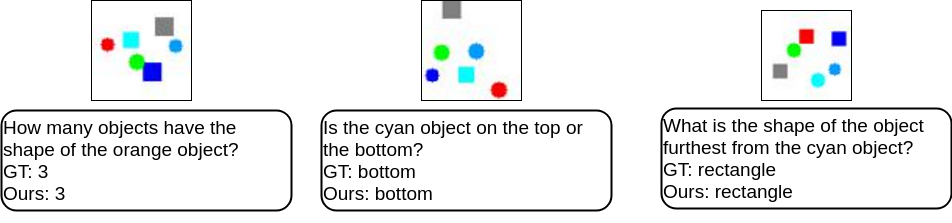}}\hfill
    \subfloat{\includegraphics[width=0.5\textwidth, height=0.1\textheight]{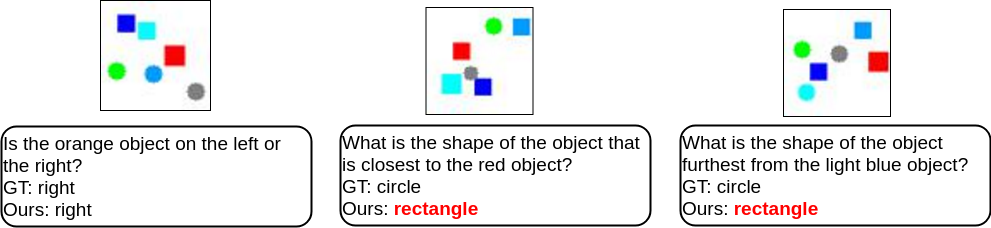}}
    \caption{Some example images, questions and answers from the synthetically generated Sort-of-Clevr dataset. Red-colored answers indicate failure cases. 
     }
\label{fig:results_soc}
\end{figure}

 {\bf Student Learning: }One may argue that adding such additional information to a model can be an unfair comparison. 
 However, in this work, our main aim is to integrate additional knowledge {\color{black}{(when it is available)}} with existing neural network architectures and demonstrate the benefits that such knowledge can provide. We experiment with the knowledge distillation paradigm to distill knowledge to a student. {\color{black}{Extracted knowledge can be noisy, imperfect and often costly at test time. The distillation paradigm helps in this regard as the student network can choose to learn from the ground-truth data (putting less weight on teacher's predictions) during the training phase and doesn't require the additional knowledge during test time.}} For Sort-of-Clevr, we see an accuracy of \textbf{88.2}$\%$ achieved by the student network (in external mask setting), whereas for CLEVR the distillation effort increases the accuracy over the baseline method by $2\%$. Lastly, we show some qualitative {\color{black}{examples of student network's output on}} the Sort-of-Clevr dataset (Fig.~\ref{fig:results_soc}). {\color{black} The qualitative results indicate that our method can handle counting, spatial relationships well, but fails mostly on cases relating to shapes.  This observation coupled with improvement in generalization validates that the spatial knowledge has a significant role in our method.}

\section{Conclusion}
  There has been a significant increase in attempts to integrate background knowledge (linguistic knowledge \cite{yu2017_vrd_knowledge_distillation} or commonsense rules \cite{hu-EtAl:2016:P16-1}) with state-of-the-art Neural Architectures in Computer Vision and Natural Language Processing applications. In this work, we attempt to integrate {\color{black}{additional information in the form of}} spatial knowledge with {\color{black}{existing}} neural networks to aid Visual Reasoning. The spatial knowledge is obtained by reasoning on the natural language question and additional scene information using the Probabilistic Soft Logic inference mechanism. We show that such information can be encoded using a mask over the image {\color{black}{and integrated with neural networks using knowledge distillation. Such a procedure}} shows {\color{black}{significant improvement}} on the accuracy over the baseline network.

\section{Acknowledgements}
The support of the National Science Foundation under the Robust Intelligence Program (1816039 and 1750082), and a gift from Verisk AI are gratefully acknowledged. We also acknowledge NVIDIA for the donation of GPUs.

\printbibliography[notkeyword=supp]

\clearpage

\thispagestyle{plain}
\begin{center}
    \Large
    \textbf{Supplementary Material: \\ Spatial Knowledge Distillation to aid Visual Reasoning}
\end{center}

\setcounter{section}{0}
\renewcommand{\thesection}{\Roman{section}} 
\section{External Mask Prediction Example} 

We first describe how we obtain the predicate confidence scores for both datasets. We use the image and the question from Fig.~\ref{archi} as the running example.

\begin{figure}[htb!]
	\centering
    \includegraphics[width=0.5\textwidth,height=0.25\textheight]{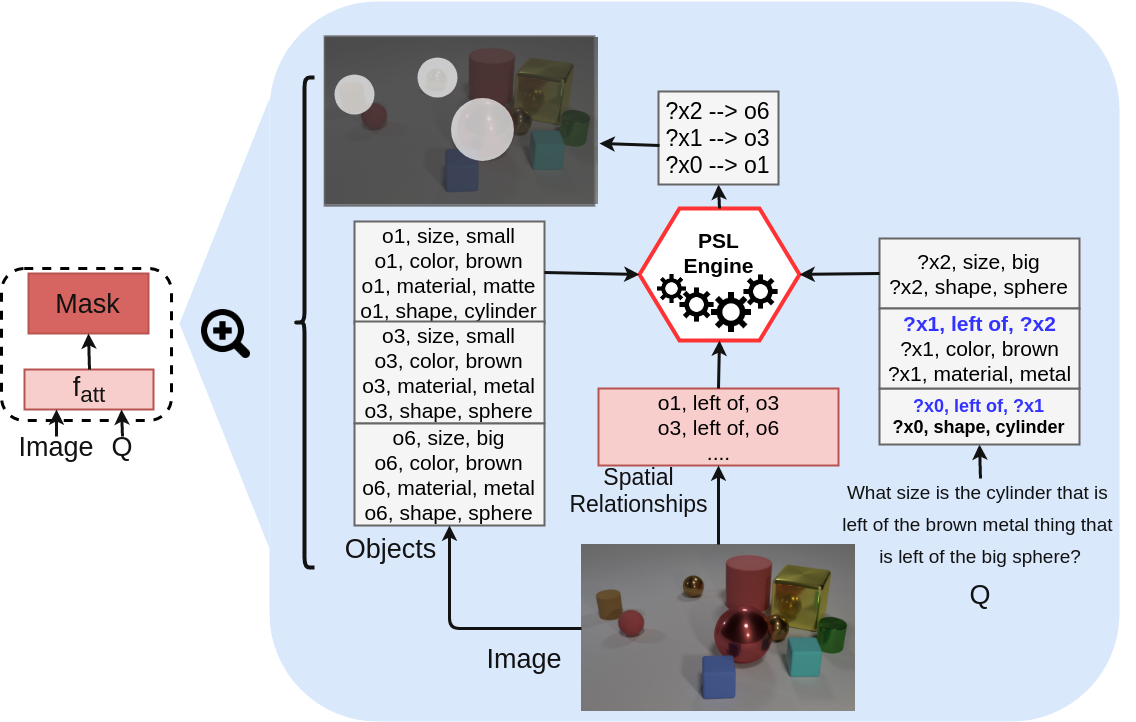}
    \caption{Internal process of mask creation. Best seen when zoomed in.}
    \label{archi}
    \vspace{-10pt}
\end{figure}

From the main paper, the two rules required to estimate which object mentions are related to which textual mentions are as follows:
\begin{equation*}
\setlength{\abovedisplayskip}{1pt}
\setlength{\belowdisplayskip}{1pt}
\begin{aligned}
w_1: candidate(M,O) &\leftarrow object(O) \land mention(M) \\&\land attr_o(O,A,V) \land attr_m(M,A,V).
\end{aligned}
\end{equation*}  
\begin{equation*}
\begin{aligned}
w_2: candidate(M,O) &\leftarrow object(O) \land mention(M) \\&\land candidate(M,O) \\&\land candidate(M_1,O_1) \\& \land consistent(A,O,O_1,M,M_1).
\end{aligned}
\end{equation*}  

$attr_o(O,A,V)$ was directly obtained by leveraging the synthetic data generation process, which is similar to CLEVR dataset generation \cite{johnson2016clevr_2}. For example $attr_o(o_1,size,small)=1.0, attr_o(o_1,$ $material,matte)=1.0$ for the leftmost brown cylinder for the image $I$. To obtain confidence scores for $attr_m(M,A,V)$, we parse the natural language question using the Stanford syntactic dependency parser \cite{de2006generating} to obtain all nouns. For all the nouns, we extract the qualifying adjectives and each qualifying adjective is assigned to an attribute (shape, size, color, material) using a similarity measure (average similarity based on Word2vec and WordNet\footnote{WordNet-based word pair similarities is calculated as a product of \texttt{length} (of the shortest path between sysnsets of the words), and \texttt{depth} (the depth of the subsumer in the hierarchical semantic net) \cite{li2006sentence}.}). For the example question, we obtain $attr_m(?x0,shape,$ $cylinder)=1.0$, $attr_m(?x1,color,brown)=1.0$. Then, for each textual mention $M$, we maintain a list of objects, where an object is only filtered out if the object and mention have a conflicting property-value pair. To obtain the $consistent(R,O,O_1,M,M_1)$ values, we perform the following steps: 1) for each mention-pair ($M$, $M_1$), we choose a corresponding candidate object-pair ($O$, $O_1$), 2) for the mention-pair we extract the shortest-path from the syntactic dependency tree and match with the type of attribute (\textit{size, shape, left, right, beside}) using the highest word-similarity measure, 3) if the attribute is a property  (such as shape, size, color), then the mentioned relation is found (\textit{same, as large as, larger than, greater than}) and the property values of objects $O$ and $O_1$ are used to check their consistency. If they are consistent we use $1.0$ or else we use $0.0$ as the score; and 4) if the attribute is spatial (such as \textit{left to, right to, beside, next}) then we check the spatial relationship and use the confidence of $1.0$ if the object-pair $O,O_1$ is consistent, otherwise we use $0.0$; for example $consistent(\text{left},o3,o6,?x1,?x2)$ $=1.0$ in the example image. Using the above predicate values, we use the PSL engine to infer the candidate objects and calculate the ground-truth mask.
\printbibliography[keyword=supp, resetnumbers=true]
\end{document}